\documentclass[letterpaper, 10pt, conference]{ieeeconf}      
 
\IEEEoverridecommandlockouts                              

\usepackage{algorithm}
\usepackage{multicol}
\usepackage{multirow}
\usepackage{algpseudocode}
\usepackage{color}
\usepackage{xcolor}
\usepackage{graphicx}
\usepackage{bm}
\usepackage{amsmath}
\usepackage{amssymb}
\usepackage{tabularx}
\usepackage{subcaption}
\usepackage{url}
\usepackage{cite}

\newcommand{\minus}{\scalebox{0.75}[1.0]{$-$}}

\usepackage{hyperref}
\hypersetup{
    colorlinks,
    linkcolor={blue!80!black},
    citecolor={blue!80!black},
    urlcolor={blue!100!black}
}

\title{\LARGE  \bf
CrowdMove: Autonomous Mapless Navigation in Crowded Scenarios} 

\author{Tingxiang Fan$^{1}$, Xinjing Cheng$^{1}$, Jia Pan$^{2}$, Dinesh Manocha$^{3}$ and Ruigang Yang$^{1}$
\thanks{$^{1}$Tingxiang Fan, Xinjing Cheng and Ruigang Yang are with Researcher of Robotics and Auto-driving Lab, Baidu Research, Baidu Inc., China
        {\tt\small v\_fantingxiang@baidu.com, chengxinjing@baidu.com, yangruigang@baidu.com}}%
\thanks{$^{2}$Jia Pan is with the Department of Mechanical and Biomedical Engineering, City University of Hong Kong, Hong Kong, China {\tt\small jiapan@cityu.edu.hk}}
\thanks{$^{3}$Dinesh Manocha is with Department of Computer Science, University of Maryland, College Park, USA
        {\tt\small dm@cs.umd.edu}}
}

\begin{document}
\maketitle

\ieeefootline{Workshop on From freezing to jostling robots: Current challenges and new paradigms for safe robot navigation in dense crowds \\ International Conference on Intelligent Robots and Systems 2018, Madrid, Spain}
\begin{abstract}

Navigation is an essential capability for mobile robots. In this paper, we propose a generalized yet effective 3M (i.e., multi-robot, multi-scenario, and multi-stage) training framework. We optimize a mapless navigation policy with a robust policy gradient algorithm. Our method enables different types of mobile platforms to navigate safely in complex and highly dynamic environments, such as pedestrian crowds. To demonstrate the superiority of our method, we test our methods with four kinds of mobile platforms in four scenarios. Videos are available at \url{https://sites.google.com/view/crowdmove}
\end{abstract}

\section{Introduction}
\label{sec:intro}

Safe and efficient navigation in highly dynamic unstructured environments remains an open problem in robotics~\cite{vemula2017modeling,bera2017sociosense}. As a result, the mobility of robots nowadays is still limited in a crowded pedestrian scenarios, which greatly limits the mobile robot's application in many tasks, including the restaurant delivery and the surveillance.

Most existing robotic navigation approaches consist of two parts. First, a map is built online/offline using simultaneous localization and mapping (SLAM). Next, a collision-free trajectory is planned with respect to the map~\cite{shen2014multi}. However, in an environment full of moving obstacles like pedestrians, SLAM may fail frequently due to occlusions or noises and may not be able to provide a reliable map. 
More importantly, a map about surrounding environment often is not necessary for a robust collision avoidance policy, since the collision avoidance is a reactive behavior that only requires a rough sensing about the position and velocity of nearby obstacles. In addition, the map construction is expensive and thus SLAM is not applicable in many mobile robotic platforms with limited computational resources.

Given the above limitations, mapless navigation approaches have been proposed recently, in which robots do not rely on the prior knowledge of surrounding environments~\cite{oleynikova2015reactive,choi2017real}. For the mapless navigation, the localization service is no longer provided by the SLAM, but by other low-cost solutions such as the Global Position System (GPS) for outdoor scenes and the Ultra Wide Band (UWB) technique for indoor scenes~\cite{ye2010high}. To develop the collision avoidance capability, some manual rules such as artificial potential field (APF)~\cite{haddad1998reactive} have been proposed. However, these hand-crafted rules is limited in generalization and the collision avoidance performance is sensitive to the rules' hyper-parameters.

Recently, deep learning based mapless navigation approaches have gained attention, which directly map sensor perception to steering commands~\cite{tai2017virtual,ort2018autonomous}. Unlike rule-based approaches, learning-based navigation methods are optimized over a large number of training data and thus no longer require the manual tuning of hyper-parameters and rules. Robot-learning schemes could be divided into two categories: the imitation learning and the reinforcement learning~\cite{abbeel2007application,levine2016end}. In imitation learning, robots learn optimal policies by imitating demonstrations collected from human experts~\cite{argall2009survey}. In reinforcement learning, the optimal policies are learned from the training data generated during the interaction between robots and the environment~\cite{mnih2015human}. 

In this paper, we focus on the second category, i.e., the reinforcement learning, to address the problem of safe and efficient navigation in crowds. In our approach, the local planner is modeled by a deep neural network, which transfers raw sensor inputs to a collision-free steering command. Thanks to its excellent generalization, the local planner trained in a simulator can be deployed in the real world without tedious fine-tuning. 

\begin{figure*}
\centering
\includegraphics[width=1\linewidth]{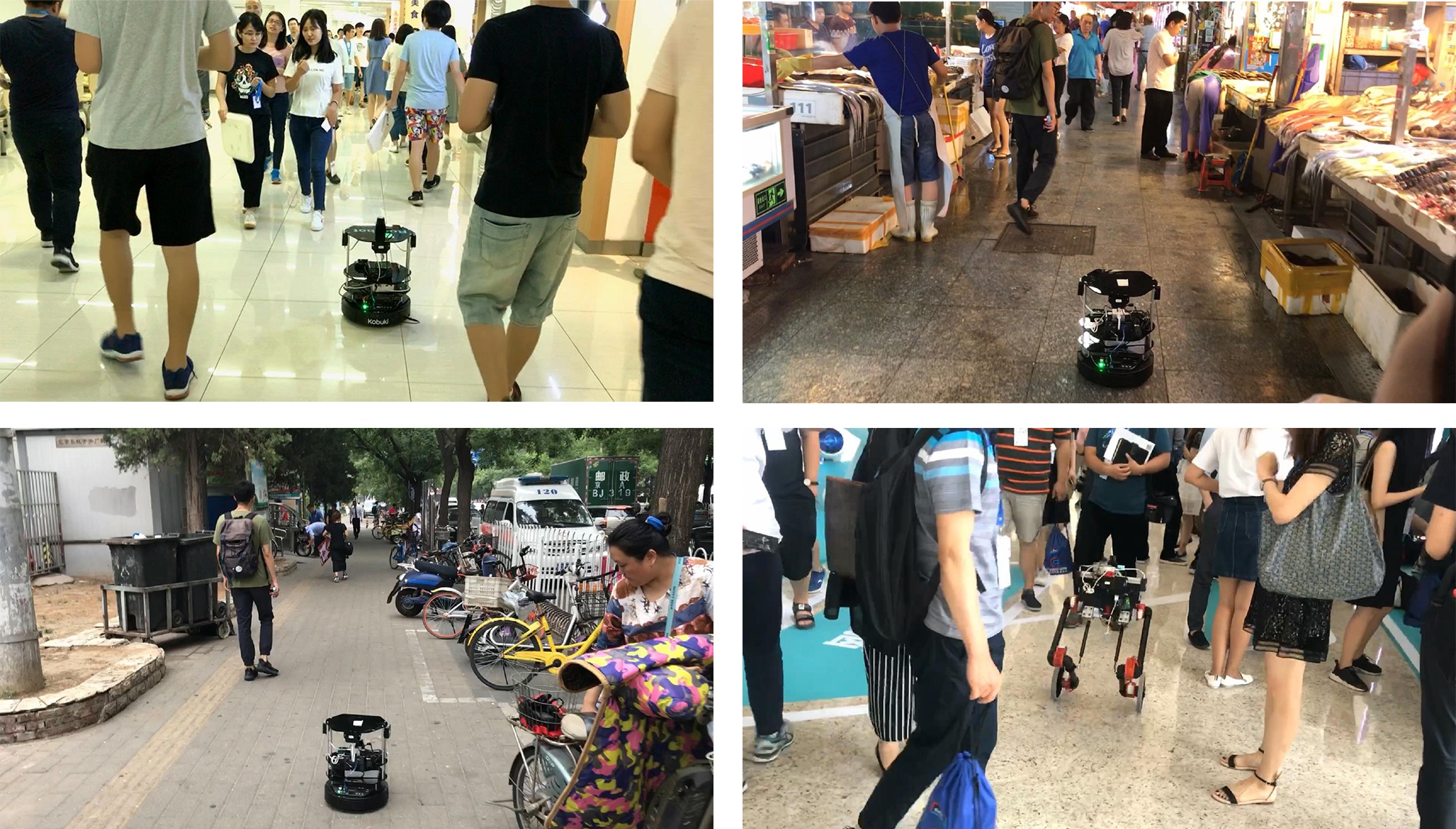} 
\caption{Mapless navigation in complex and highly dynamic environments using different mobile platforms.}
\label{fig:cover}
\end{figure*}

\noindent \textbf{Main contributions:} 
To learn a robust mapless local planner, we propose a novel 3M (i.e., multi-robot, multi-scenario, and multi-stage) training framework, which exploits a robust policy
gradient based reinforcement learning algorithm that is trained
in a large-scale robot system in a set of complex environments.
We demonstrate that the collision avoidance policy learned from our approach can find collision-free paths for nonholonomic robots and can be generalized effectively to various scenarios and different mobile platforms, as shown in Figure~\ref{fig:cover}. 

\section{Related Works}
\label{sec:related work}

Collision avoidance is an essential capability for mobile robotic systems. Standard rule-based solutions to the collision-free navigation have two steps: first, the environment is modeled as an energy function, and then a collision-free path is computed as an optimization problem for reaching the destination~\cite{siciliano2016springer}. These approaches benefit from the independent development of mapping-based localization and the motion planning~\cite{shen2013vision, mohta2014vision}. The major limitation of rule-based methods is that the energy functions are always hand-crafted, meaning that they need a lot of expertise and may fail in complex and highly dynamic scenarios.

Learning-based collision avoidance techniques in which one robot avoids static obstacles have been studied extensively. Many approaches adopt the supervised learning paradigm to train a collision avoidance policy by imitating a dataset of sensor input and motion commands.
Muller et al.~\cite{muller2006off} trained a vision-based static obstacle avoidance system in supervised mode for a mobile robot by training a 6-layer convolutional network which maps raw input images to steering angles. 
Sergeant et al.~\cite{sergeant2015multimodal} proposed a mobile robot control system based on multimodal deep autoencoders. 
Ross et al.~\cite{ross2013learning} trained a discrete controller for a small quadrotor helicopter with imitation learning techniques. The quadrotor was able to successfully avoid collisions with static obstacles in the environment using only a single cheap camera. Only discrete movements (left/right) must be learned and the robot need only be trained within static obstacles. 
Note that the aforementioned approaches only take into account static obstacles and require a human driver to collect training data in a wide variety of environments. 
Another data-driven end-to-end motion planner is presented by Pfeiffer et al.~\cite{pfeiffer2017perception}. They trained a model to map laser range findings and target positions to motion commands using expert demonstrations generated by the ROS navigation package. This model can navigate the robot through a previously unseen environment and ensure that it successfully reacts to sudden changes. Nonetheless, like the other supervised learning methods, the performance of the learned policy is severely constrained by the quality of the labeled training sets. 
To overcome this limitation, Tai et al.~\cite{tai2017virtual} proposed a mapless motion planner trained through a deep reinforcement learning method. Kahn et al.~\cite{kahn2017uncertainty} presented an uncertainty-aware model-based reinforcement learning algorithm to estimate the probability of collision in an priori unknown environment. However, the test environments are relatively simple and structured, and the learned planner is hard to generalize to scenarios with dynamic obstacles and other proactive agents. To address highly dynamic unstructured environments, some researches which applied decentralized multi-robot navigation approaches, are proposed recently.  
Godoy et al.~\cite{godoy2016moving} introduced Bayesian inference approach to predict surrounding dynamic obstacles and computed collision-free command through ORCA framework~\cite{van2011reciprocal}. Chen et al.~\cite{chen2017decentralized,chen2017socially,everett2018motion} proposed multi-robot collision avoidance policies by deep reinforcement learning, which also required deploying multiple sensors to estimate the states of nearby agents and moving obstacles. However, complex pipeline of these navigation approaches not only requires expensive online computation but makes the whole system less robust to the perception uncertainty.
\section{Approach}
\label{sec:approach}
The approach in this paper stems from our previous work~\cite{long2017towards}. Here we provide a brief description of this approach in terms of the key ingredients of our reinforcement learning framework, network structure and the training procedure.

\subsection{Reinforcement Learning Setup}
\label{sec:setup}
In reinforcement learning, the environment is typically formulated as a Markov Decision Process(MDP) which can be described as a 4-tuple $(\mathcal{S}, \mathcal{A}, \mathcal{P}, \mathcal{R})$, where $\mathcal{S}$ is the state space, $\mathcal{A}$ is the action space, $\mathcal{P}$ is the state-transition model, $\mathcal{R}$ is the reward function. Below we describe the details of the state space, the action space, and the reward function. 

\subsubsection{\textbf{State space}} 
The state $\mathbf{s}^t$ consists of the readings of the 2D laser range finder $\mathbf{s}_z^t$, the relative goal position $\mathbf{s}_g^t$ and robot's current velocity $\mathbf{s}_v^t$. Specifically, $\mathbf{s}_z^t$ includes the measurements of the last three consecutive frames from a 180-degree laser scanner which has a maximum range of 4 meters and provides 512 distance values per scanning (i.e. $\mathbf{s}_z^t \in \mathbb{R}^{3 \times 512}$). The relative goal position $\mathbf{s}_g^t$ is a 2D vector representing the goal in polar coordinate (distance and angle) with respect to the robot's current position. The observed velocity $\mathbf{s}_v^t$ includes the current translational and rotational velocity of the differential-driven robot. The observations are normalized by subtracting the mean and dividing by the standard deviation using the statistics aggregated over the course of the entire training.  
\subsubsection{\textbf{Action space}} 
The action space is a set of permissible velocities in continuous space. The action of differential robot includes the translational and rotational velocity, i.e. $\mathbf{a}^{t} = [v^{t}, w^{t}]$. In this work, considering the real robot's kinematics and the real world applications, we set the range of the translational velocity $v \in (0.0, 1.0)$ and the rotational velocity in $w \in (-1.0, 1.0)$. Note that backward moving (i.e. $v < 0.0$) is not allowed since the laser range finder cannot cover the back area of the robot. 
\subsubsection{\textbf{Reward design}} 
Our objective is to avoid collisions during navigation and minimize the arrival time. A reward function is designed to guide robots to achieve this objective: 
\begin{equation}
r^t = (^gr)^t + (^cr)^t + (^wr)^t. 
\end{equation}
The reward $r$ received at timestep $t$ is a sum of three terms, $^gr$, $^cr$, and $^wr$. In particular, the robot is awarded by $(^gr)^t$ for reaching its goal:
\begin{equation} 
(^gr)^t = 
\begin{cases}
  r_{arrival}  & \ \text{if } \| \mathbf{p}^t - \mathbf{g} \| < 0.1 \\
  \omega_g(\|\mathbf{p}^{t-1}- \mathbf{g} \| - \|\mathbf{p}^{t}- \mathbf{g} \|) & \ \text{otherwise}. \\
\end{cases}
\end{equation}
When the robot collides with other robots or obstacles in the environment, it is penalized by $(^cr)^t$: 
\begin{equation}
(^cr)^t =  
\begin{cases}
r_{collision} & \quad \text{if } collision \\
0 & \quad \text{otherwise}.  \\
\end{cases}
\end{equation}
To encourage the robot to move smoothly, a small penalty $(^wr)^t$ is introduced to punish the large rotational velocities: 
\begin{equation}  
(^wr)^t = \omega_{w}|w^t|  \quad \quad \text{if }  |w^t| > 0.7 . 
\end{equation}
We set $r_{arrival}=15$, $\omega_g=2.5$, $r_{collision}=\minus 15$ and $\omega_w=\minus 0.1$ in the training procedure.

\subsection{Network architecture}
\label{sec:model}

\begin{figure}
\centering 
\includegraphics[width=0.9\linewidth]{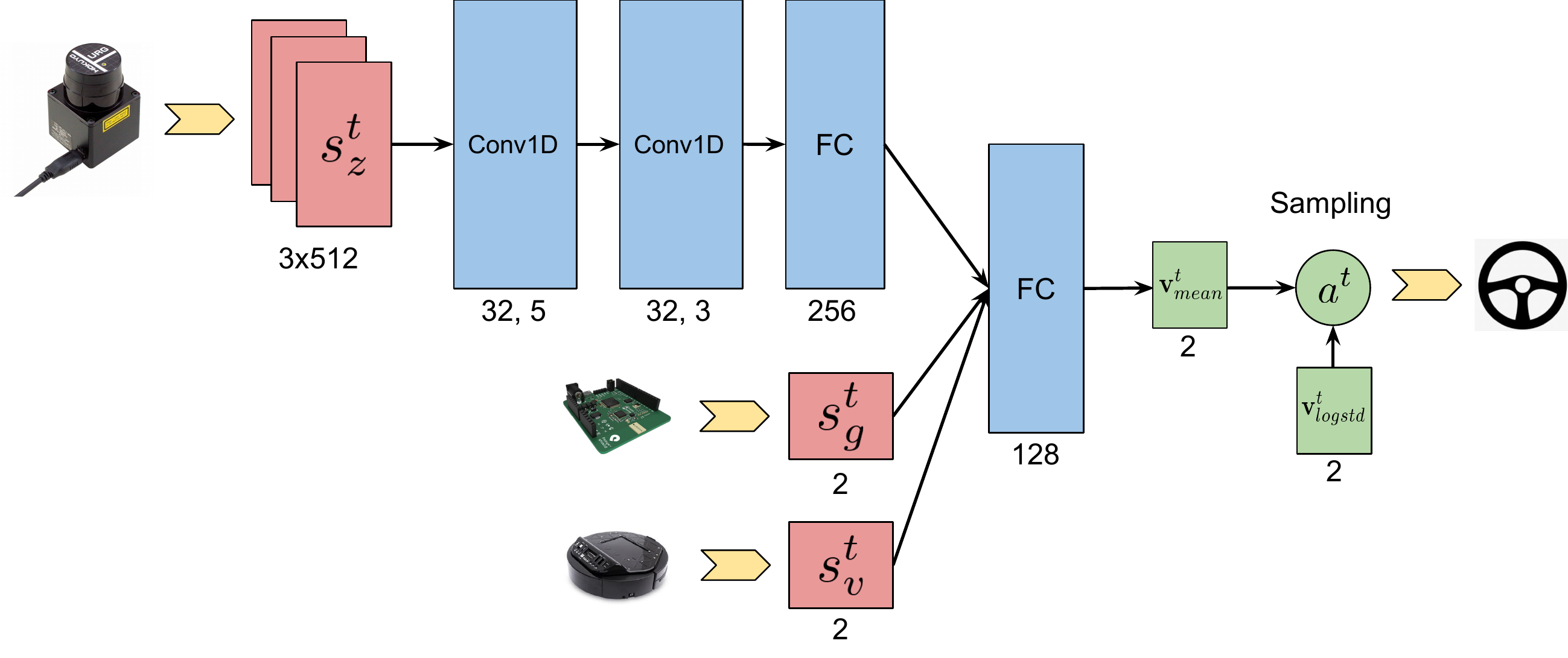} 
\caption{The architecture of the collision avoidance neural network. The network has the scan measurements $\mathbf{s}^t_z$, relative goal position $\mathbf{s}^t_g$ and current velocity $\mathbf{s}^t_v$ as inputs, and outputs the mean velocity $\mathbf{v}^t_{mean}$. } 
\label{fig:model} 
\end{figure}

We design a 4-hidden-layer neural network as a non-linear function approximator to the policy $\pi_{\theta}$. Its architecture is shown in Figure~\ref{fig:model}. The neural network maps the input state vector $\mathbf{o}^t$ to a vector $\mathbf{v}_{mean}^t$. 
The final action $\mathbf{a}^t$ is sampled from a Gaussian distribution $\mathcal{N}(\mathbf{v}^t_{mean}, \mathbf{v}_{logstd}^t)$,
where $\mathbf{v}^t_{mean}$ serves as the mean and $\mathbf{v}_{logstd}^t$ refers to a log standard deviation which will be updated solely during training.

\subsection{Multi-robot, multi-scenario, and multi-stage training}
\subsubsection{Training algorithm} 
\label{sec:train}
We extend the state-of-the-art reinforcement learning algorithm, Proximal Policy Optimization (PPO)~\cite{schulman2017proximal,heess2017emergence,trpo}, to our parallel multi-robot training framework. The policy is trained with experiences collected by all robots in parallel. The parallel multi-robot training framework not only dramatically reduces the time cost of the sample collection but also makes the algorithm suitable for training many robots in various scenarios. 

\subsubsection{Training scenarios} 
\begin{figure}[t] 
\centering
\includegraphics[width=1\linewidth]{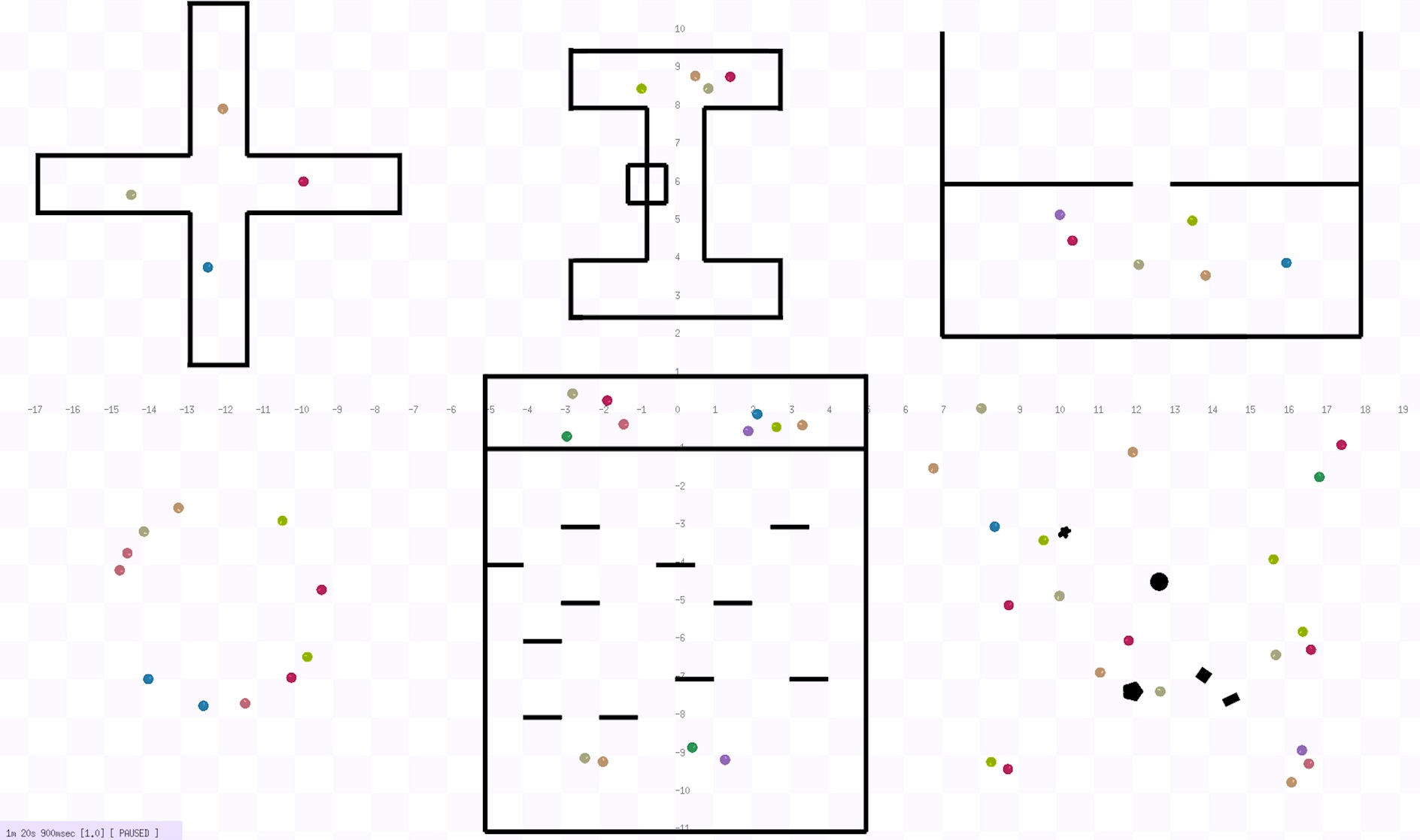}
\caption{Scenarios used to train the collision avoidance policy. All robots are modeled as discs with the same radii. Obstacles are shown in black.}
\label{fig:scene}
\end{figure}

To expose our robots to diverse environments, we create different scenarios with a variety of obstacles using the Stage mobile robot simulator\footnote{\url{http://rtv.github.io/Stage/}} (as shown in Figure~\ref{fig:scene}) and move all the robots concurrently.  These rich, complex training scenarios enable robots to explore state space efficiently and are likely to improve the quality and robustness of the learned policy. Combined with the parallel \textit{multi-robot} training algorithm, the collision avoidance policy is effectively optimized at each iteration over a variety of environments. 

\subsubsection{Training stages} 
\begin{figure} 
\centering
\includegraphics[width=1\linewidth]{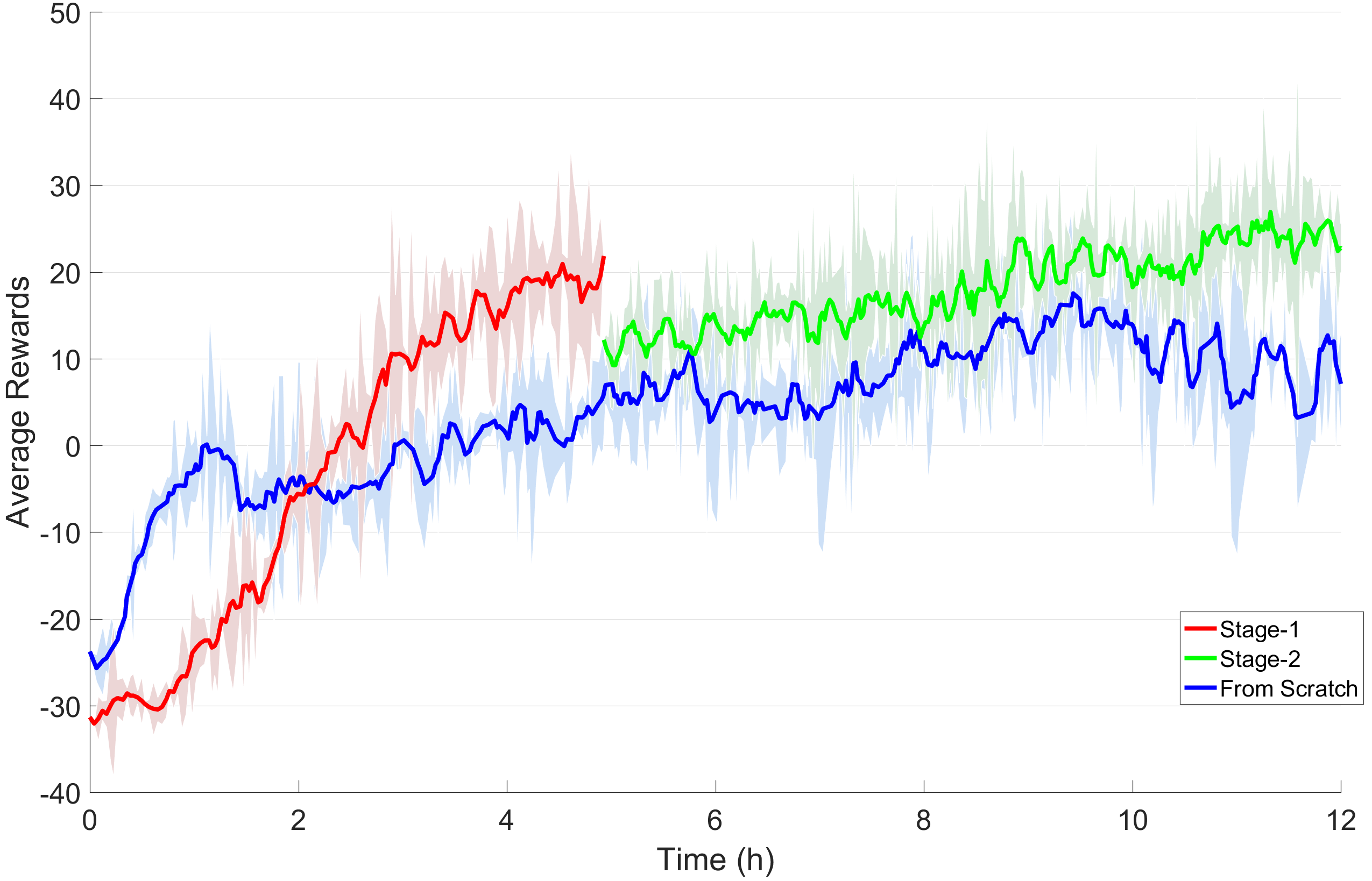}
\caption{Average rewards shown in wall time during the training process.}
\label{fig:reward}
\end{figure}

Although training in multiple environments simultaneously enables robust performance in a variety of different test cases, it makes the training process harder. Inspired by the curriculum learning paradigm~\cite{bengio2009curriculum}, we propose a two-stage training process, which accelerates the policy's convergence to a satisfying solution, and gets higher rewards than if the policy had been trained from scratch with the same number of epochs (as shown in Figure~\ref{fig:reward}). In the first stage, we only train 20 robots on the random scenarios (scenario 7 in Figure~\ref{fig:scene}) without any obstacles. Once the robots achieve reliable levels of performance, we stop Stage 1 and save the trained policy. The policy will continue to be updated in Stage 2, where the number of robots increases to 58, and they are trained on the richer and more complex scenarios shown in Figure~\ref{fig:scene}. 

\section{Experiments and Results}
\label{sec:exp}

In this section, we first describe the hardware setup of our robots. Then, we deploy our learning-based local planner to a Turtlebot platform and test the performance in various crowd scenarios. Finally, to validate the transferability of our model, we apply the collision avoidance policy on different mobile platforms. 

\subsection{Hardware setup}
We now introduce the sensor kits and the mobile platforms used in our experiments. The sensor kits provide the input to the collision avoidance network, and the mobile platforms execute the steering command output from the collision avoidance network.

\subsubsection{Sensor kits}
We used the Hokuyo urg-04lx 2D LiDAR as our laser scanner, the Pozyx Ultra-WideBand(UWB) modules as the indoor localization system, and the Nvidia Jetson TX1 as our onboard computer. In this way, the cost of the sensor kit is economical.

\subsubsection{Mobile platforms}
\begin{figure}
\captionsetup[subfigure]{justification=centering}
\centering
\begin{subfigure}{0.22\textwidth}
\centering
\includegraphics[width=.8\linewidth]{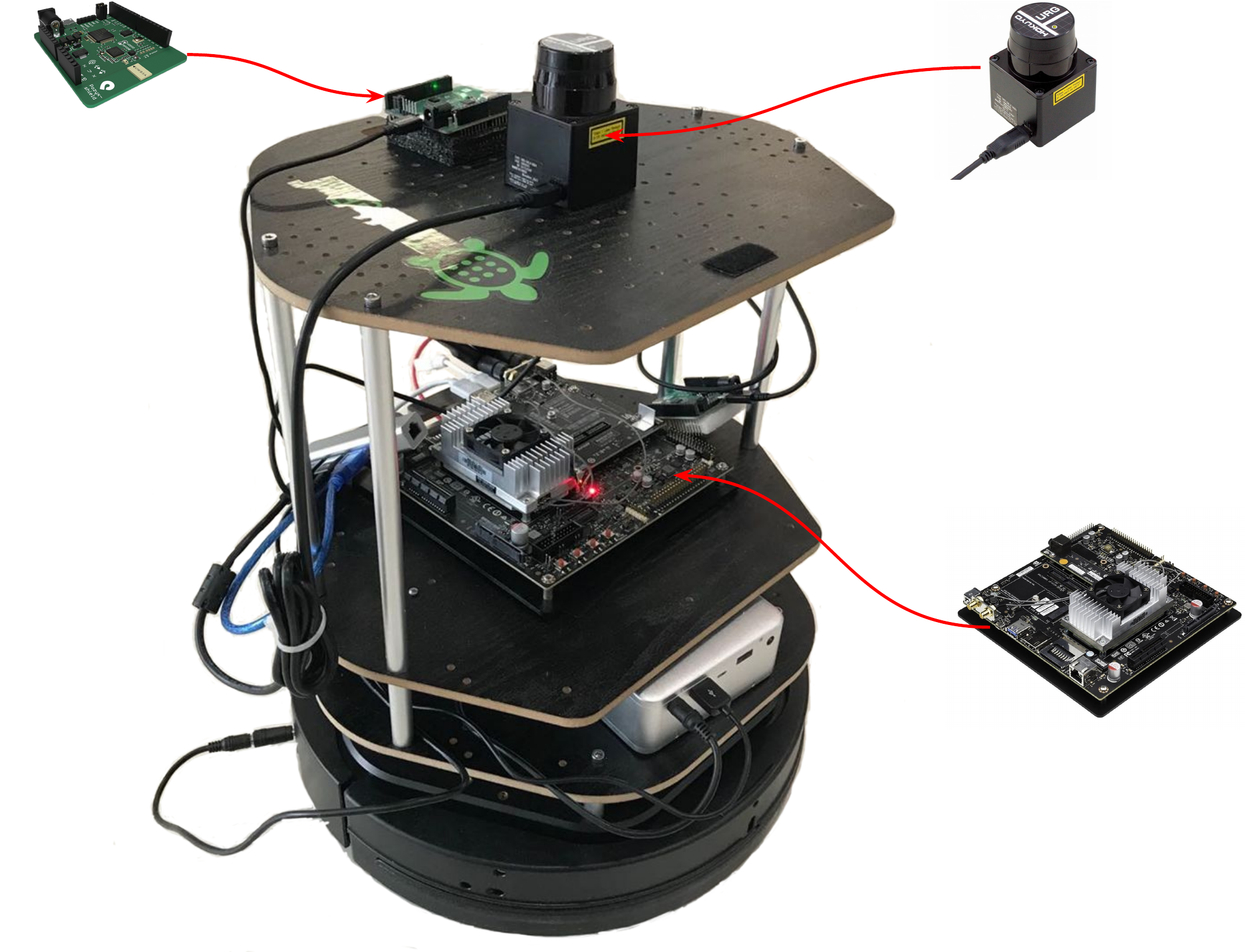}
\caption{Turtlebot}
\label{fig:hardware_turtlebot}
\end{subfigure}
\begin{subfigure}{0.22\textwidth}
\centering
\includegraphics[width=0.43\linewidth]{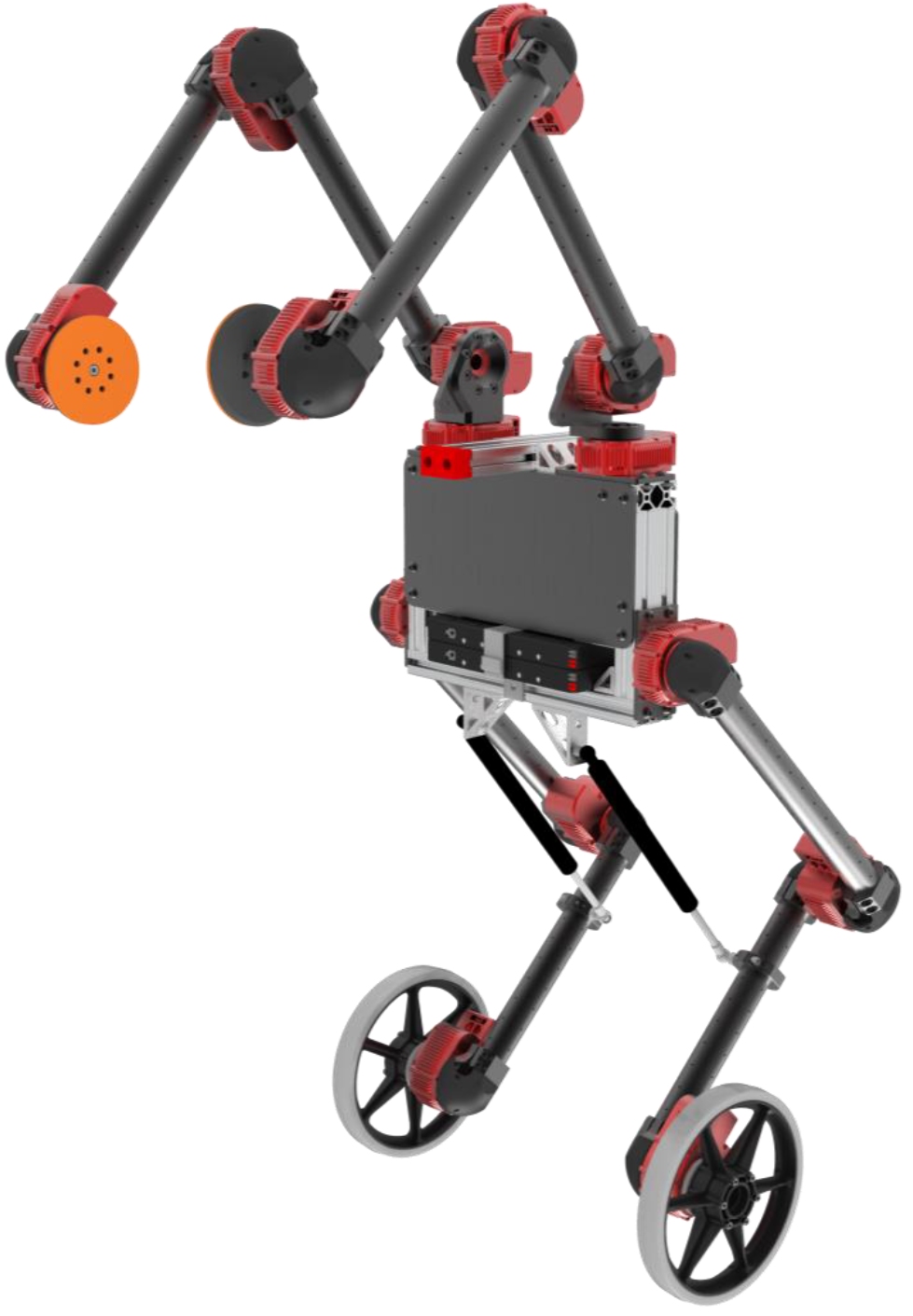}
\caption{Igor robot}
\label{fig:hardware_igor}
\end{subfigure}
\\
\centering
\begin{subfigure}{0.22\textwidth}
\centering
\includegraphics[width=0.24\linewidth]{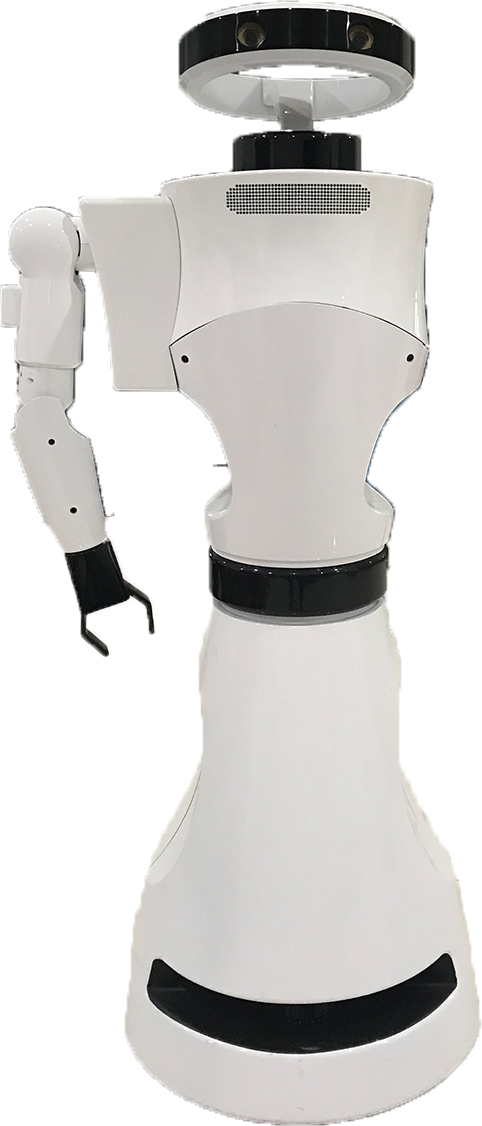}
\caption{Human-like service robot}
\label{fig:hardware_baidubear}
\end{subfigure}
\begin{subfigure}{0.22\textwidth}
\centering
\includegraphics[width=0.34\linewidth]{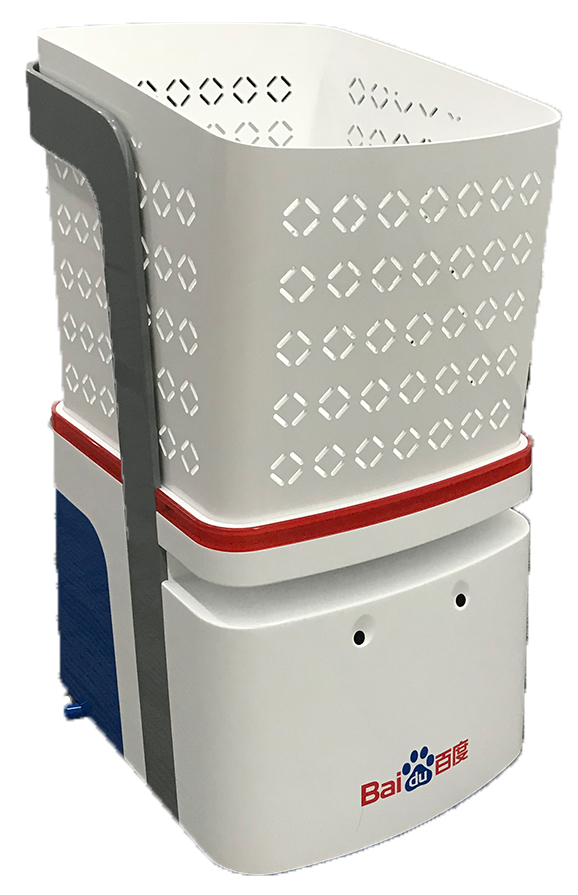}
\caption{Shopping cart}
\label{fig:hardware_cart}
\end{subfigure}
\caption{Four different mobile robots used in our experiments. (a) shows
the sensor kit installed in the Turtlebot. (b)(c)(d) show another three types of mobile platforms boarded with the same collision avoidance algorithm.}
\label{fig:hardware_different}
\end{figure}

We expect our policy to have a potential applicability to different mobile robots without being retrained. To analyze the transferability of our local planner from the simulation to the real-world, four different mobile platforms have been tested in the experiments (as shown in Figure~\ref{fig:hardware_different}). 

In our experiments, to force the robot to encounter pedestrians, we let a person take the UWB localization tag, and then the robot can follow the target person according to the UWB signal. In this case, the target person can control the robot's moving directions and guide it to run through a dense pedestrian crowd.

\subsection{Tests in different scenarios}
We deployed our local planner on the Turtlebot and tested it in different environments to test the robustness of our collision avoidance policy in different scenarios. In particular, we let Turtlebot run in the canteen, the food market and the outdoor street because these environments provided numerous obstacles that never appear in simulation. In addition, the turtlebot (Figure~\ref{fig:hardware_turtlebot}) is larger than the robot for which we trained the model in the simulation.

\begin{figure*}
\centering
\includegraphics[width=1\linewidth]{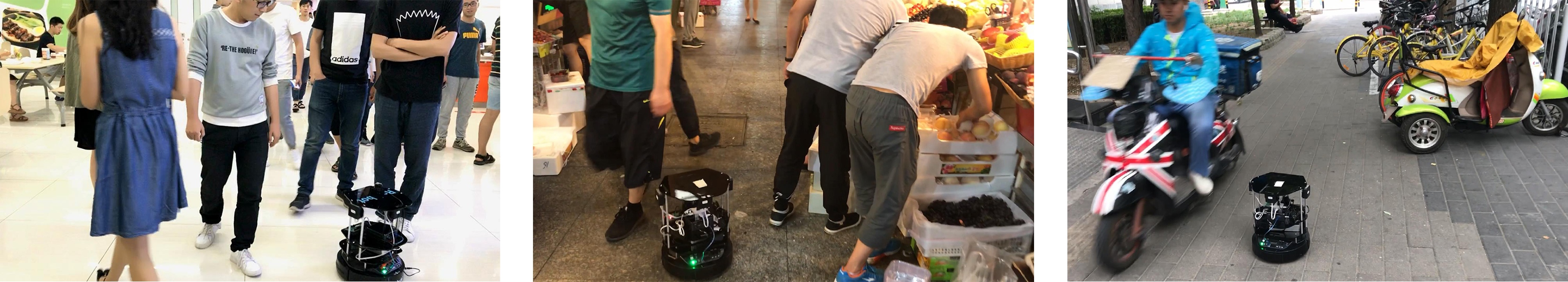}
\caption{Turtlebot running in highly dynamic unstructured scenarios.}
\label{fig:multi_scenarios}
\end{figure*}

The experiments demonstrate that the Turtlebot can always avoid pedestrians and other static obstacles (Figure~\ref{fig:multi_scenarios}), although many people are curious about the robot and actively block the robot.

\subsection{Tests on different robots}
In testing different robots, we deployed our collision avoidance policy on different mobile platforms without retraining and fine-tuning. Noted that these robots have different shapes, sizes and dynamics, but our learning-based policy can adapt well to these differences.

\begin{figure*}
\includegraphics[width=1\linewidth]{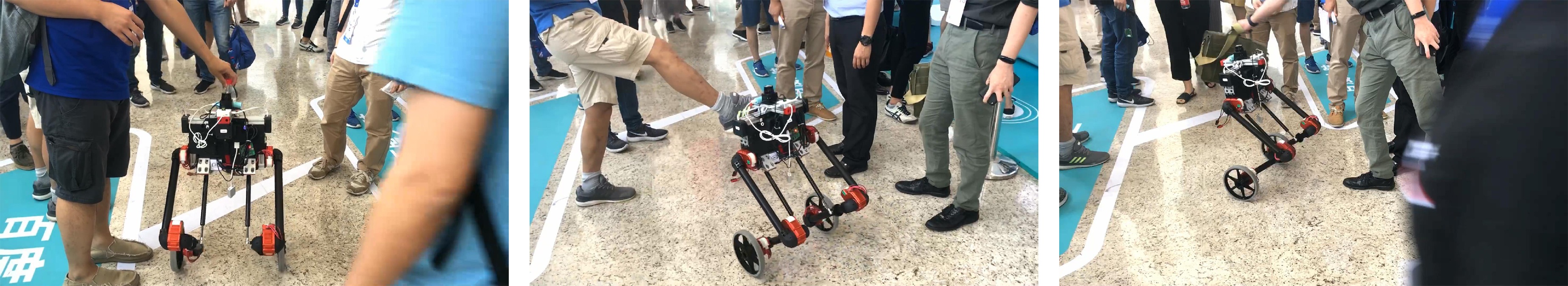}
\caption{Igor the robot reacts quickly to the bottles, human leges, and bags that suddenly appear in its field of view.}
\label{fig:igor_block}
\end{figure*}

\begin{figure*}
\centering
\includegraphics[width=1\linewidth]{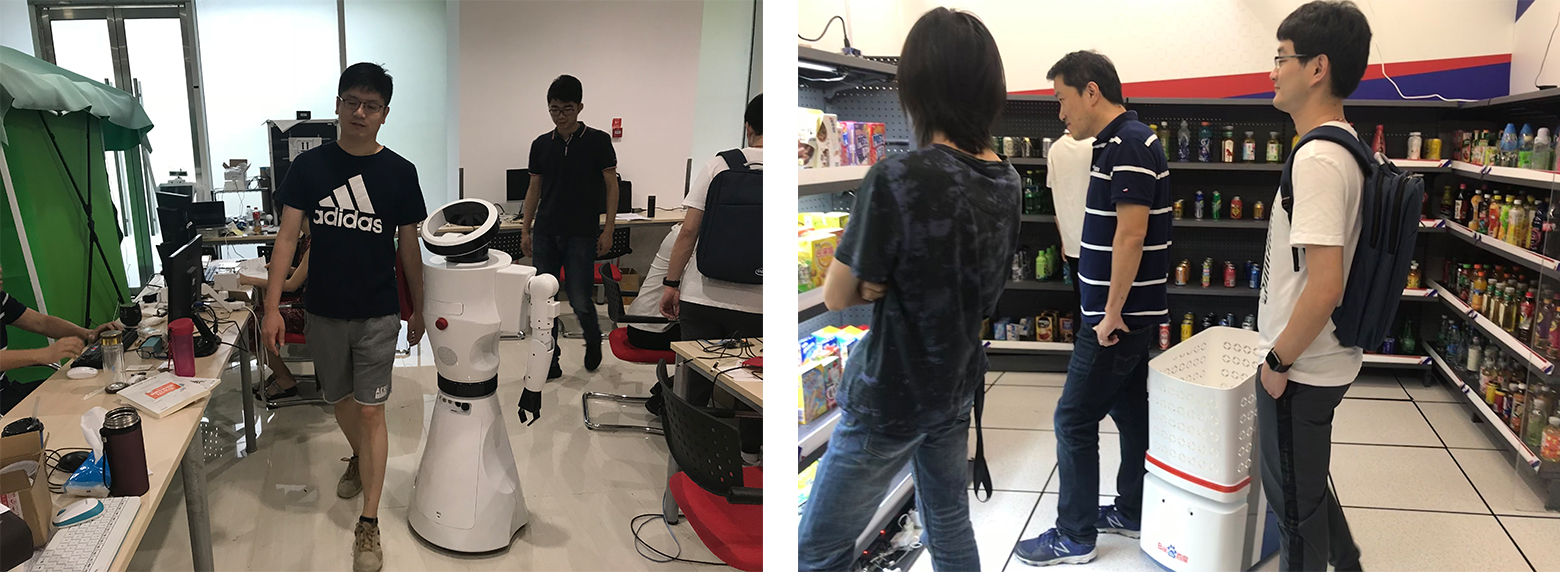}
\caption{Our algorithm was implemented on a human-like service robot and an autonomous shopping cart.}
\label{fig:multi_robot}
\end{figure*}

The experiments demonstrate that, although there are huge differences between the physcial mobile platforms and the robots in simulation, the physical robots can still avoid the static obstacles and dynamic pedestrians reliably, as shown in Figure~\ref{fig:igor_block} and~\ref{fig:multi_robot}.

\section{Conclusion}
\label{sec:conclusion}

This work presents a multi-robot, multi-scenario, and multi-stage training framework to optimize a mapless navigation policy with a robust policy gradient algorithm in simulation. The experiments demonstrate that the mapless navigation policy can achieve autonomous navigation for different mobile platforms in a large variety of crowd scenes with moving pedestrians.

{\small
\bibliographystyle{IEEEtran}
\bibliography{references}
}

\end{document}